\documentclass{article} % For LaTeX2e
\usepackage{nips15submit_e,times}
\usepackage{hyperref}
\usepackage{url}
\usepackage{amsmath}
\usepackage{graphicx}
\usepackage[ruled,linesnumbered]{algorithm2e}
\usepackage{amssymb}
\usepackage{algorithmic}
\usepackage{multirow}

\title{Group Sparse Coding for Image Denoising}

\author{
Luoyu Chen$^\dag$\\
School of Computer Engineering\\
The Australian National University\\
Canberra 2601 \\
\texttt{luoyu.chen.mlcv@gmail.com} \\
\and
Fei Wu$^\dag$\\
School of Computer Engineering\\
The Australian National University\\
Canberra 2601 \\
\texttt{wufei.mlcv@gmail.com} \\
}

\nipsfinalcopy % Uncomment for camera-ready version
\footnotetext{These authors contributed equally to this work}
\begin{document}

\maketitle
\begin{abstract}
Group sparse representation has shown promising results in image debulrring and image inpainting in GSR [3] , the main reason that lead to the success is by exploiting Sparsity and Nonlocal self-similarity (NSS) between patches on natural images, and solve a regularized optimization problem. However, directly adapting GSR[3] in image denoising yield very unstable and non-satisfactory results, to overcome these issues, this paper proposes a progressive image denoising algorithm that successfully adapt GSR [3] model  and experiments shows the superior performance than some of the state-of-the-art methods. 
\end{abstract}

\section{Introduction}
Image denoising as a classical problem has been studied in past century due to its practical significance, aiming at reconstruct high quality image from its degraded noisy version. The problem is formulated as solving a linear inverse problem: given observed noisy image $\mathbf{y}$ with additive white Gaussian noise $\mathbf{v}$ with standard deviation $\sigma$, we want to recover the original image $\mathbf{x}$, which is:$$\mathbf{y}=\mathbf{x}+\mathbf{v},\;\;\; \mathbf{v}\in \mathcal{N}(0, \sigma).$$
For most natural images, it is easy to observe there are many repeating patterns with low texture, thus, sparse coding would be ideal to represent low texture region. Simultaneously, it is hard for sparse coding to capture dense texture such as white noise, therefore appropriate level of sparse coding can capture desirable texture of original image while removing the white noise. Such idea has been explored by Ksvd~\cite{elad2006image}, a sparse coding denoising method, by iterating sparse coding and dictionary learning, more sparse representation and higher quality image can be reconstructed. However, this requires to solve a large scale optimization problem on a global dictionary, which is highly computational demanding. 
In addition to sparse coding, BM3D ~\cite{dabov2007image} also exploit NSS as an effective prior for natural images, and is one of the state of the art image denoiser. Similar patches are stacked into 3D group, and then uses two stage collaborative filtering on  transformed domain of the 3D group, which exploit both sparsity and NSS. However, its performance tend to get poorer when noise level gets high, because the patch matching phase is directly performed on noisy image.

GSR ~\cite{zhang2014group} is an improved version of Ksvd, changing unit from patch to patch group and learn adaptive dictionaries for each group, which helps prevent the large scale sparse coding for the entire image by one pass, and also acknowledge that a global dictionary is far from enough to capture rich texture features on different regions of an image, adaptive dictionaries helps yielding promising reconstruction quality. However, GSR suffers from the instability in hyper-parameter choosing to determine the trade-off between data fidelity and representation sparsity, for different images and noise levels, the optimal hyper-parameter differs significantly, which dramatically limits the possibility for large-scale batch processing. Hence, this paper's contribution focus on adapting GSR [3] to image denoising to yield stable and superior denoisng result, also, proposes a better patch similarity search scheme to further boost denoising quality.

This paper is organized as follows: first introduces the $l_0$ optimization problem for image denoising. Second, discusses the iterative optimization steps and finalize with an overview of the entire algorithm. Third part presents the experimental results.

\section{Group Sparse Representation Learning}
\subsection{Similar Patch Grouping}
In Figure 1, for an image $\mathbf{x}$, with size $N$ is divided into $n$ overlapped patches $\mathbf{x_k}$ of size $\sqrt{b}\times \sqrt{b}, k = 1,2,\cdots n$,  Then for each patch $\mathbf{x_k}$, its most similar $c$ patches are selected
from an $L \times L$ sized searching window to form a set $S_{\mathbf{x_k}}$.  After this, all the patches in $S_{\mathbf{x_k}}$ are stacked
into a matrix $\mathbf{x_{G_k}}\in \mathfrak{R}^{b\times c}$, which contains every element of $S_{\mathbf{x_k}}$ as its column, $\mathbf{x_{G_k}} = \{\mathbf{x_{{G_k},1}, x_{{G_k},2}, \cdots, x_{{G_k},c}}\}$, where $x_{{G_k},i}$ denotes the $i$-th similar patch (column form) of the $k$-th group. The similarity measurement simply uses the mean square error to determine the distance between different patches. 
% \begin{figure}[!ht]
%       \centerline{\includegraphics[width=1\linewidth]{patch_matching.png}}
%   \end{figure}
  
  \begin{figure}[h]
    \centering
    \includegraphics[width=1\linewidth]{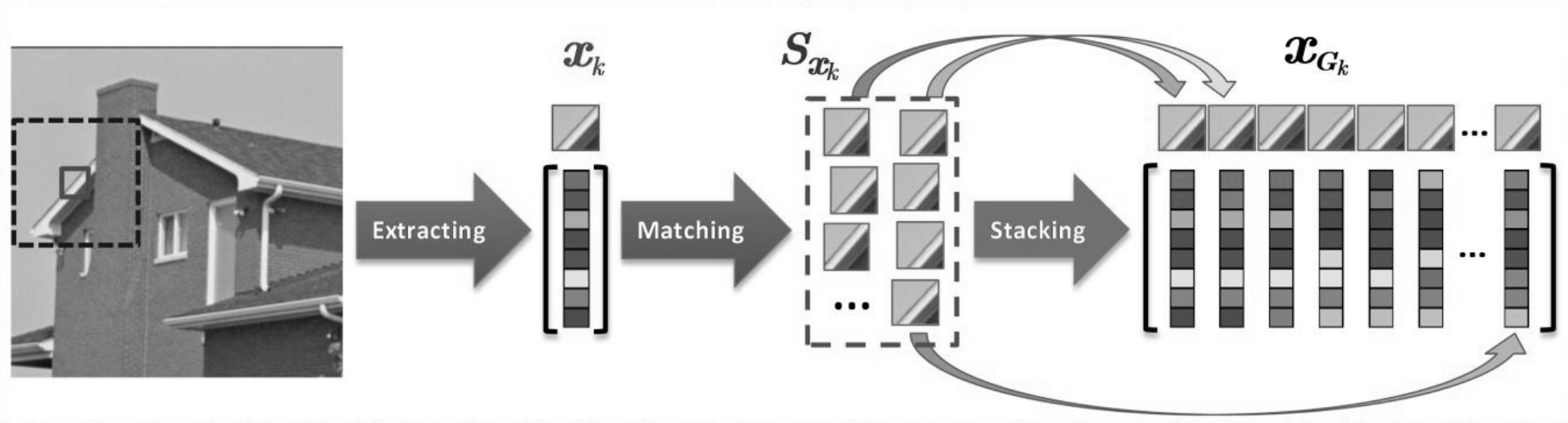}
    \caption{patch matching scheme, from GSR[3] paper}
    
    \label{fig:mesh2}
\end{figure}
In this paper, the similarity measurement is based on more and more better denoised images, therefore producing better results than purely search on the noisy image, related experiments is displayed at the third section.

\subsection{The $\mathit{l_0}$ Minimization Problem for Image Denoising }

Each group $\mathbf{x_{G_k}}$ can be sparsely represented by a dictionary $\mathbf{D_{G_k}}$ and a coefficient vector $\mathbf{a_{G_k}}$. We want to make sure that the dictionary can partially fit the noise image to capture the local texture, and also, the coefficient vector can be sparse enough to filtered out the white noise. Hence, we want to solve the optimization problem:$$\underset{\mathbf{D_x},\; \mathbf{a_x}}{\arg\min}\Sigma_{k=1}^n \|\mathbf{x_{G_k}} - \mathbf{D_{G_k}}\mathbf{a_{G_k}}\|_2 + \lambda\|\mathbf{a_x}\|_0$$
Where $\mathbf{D_x} = \{\mathbf{D_{G_1}}, \cdots,\mathbf{D_{G_n}}\}$, $\mathbf{a_x} = \{\mathbf{a_{G_1}}, \cdots,\mathbf{\mathbf{a_{G_n}}}\}$, $\lambda$ is the trade-off between representation sparsity and data fidelity, larger $\lambda$ encourages more sparse representation while sacrificing the data fidelity, which is more likely to filter out the white noise, but more sparsity generally means the recovered patches are more smooth and more local texture might be lost. Therefore, a carefully selected $\lambda$ can remove the noise while preserving most local texture.

In GSR[3], the optimization problem is solved by iteratively solving 2 sub-problems: $$\mathbf{D_x^{(t+1)}} = \underset{\mathbf{D_x}}{\arg\min}\Sigma_{k=1}^n \|\mathbf{x_{G_k}} - \mathbf{D_{G_k}}\mathbf{a_{G_k}}\|_2 + \lambda\|\mathbf{a_x}\|_0$$
$$\mathbf{a_x^{(t+1)}} = \underset{\mathbf{a_x}}{\arg\min}\Sigma_{k=1}^n \|\mathbf{x_{G_k}} - \mathbf{D_{G_k}^{(t+1)}}\mathbf{a_{G_k}}\|_2 + \lambda\|\mathbf{a_x}\|_0$$

The first sub-problem is to learn a set of adaptive dictionaries for each group, GSR[3] uses SVD decomposition for each group $\{\mathbf{x_{G_1}}, \cdots, \mathbf{x_{G_n}}\}$, the k-th group $\mathbf{x_{G_k}} = \mathbf{U_kS_kV_k^T}$, then for each local dictionary $\mathbf{D_{G_k}} = [\mathbf{u_{k,1}v_{k,1}^T},\cdots, \mathbf{u_{k,m}v_{k,m}^T}]$, where m = $\min(b, c)$.

The second sub-problem is solved by Split Bregman Iteration (SBI), the answer is approximated as, $\mathbf{a_x} = Hard(\mathbf{a_x}, 2\sqrt{\tau})$, where $\tau = (\lambda K)/(\mu N)$, $K = b\times c\times n$, $N$ is the number of pixels of image $\mathbf{x}$.  The hyper-parameter $\lambda$ and $\mu$ are empirically selected, $\lambda$ is the trade-off term as mentioned above. However, $\lambda$ varies a lot for different images and different noise level in order to obtain the best denoise effect, which means SBI approximated solution can be sub-optimal. However there still exists a 'golden-ratio' trade-off factor between texture fitting (the data fidelity term) and denoising effect (the sparsity regularization term). By experimenting on multiple images, when the mean absolute difference between reconstructed image and the current denoising image is $0.73\hat{\sigma}^{(t)}$, or the difference stops increasing when more sparse $\mathbf{a_{G_k}}$ is being thresholded (only happens when $\sigma\geq75$), the best threshoulding is obtained. $\mu$ is the learning rate to produce a less noisy image in each iteration by superimposing the currently denoised images $\hat{\mathbf{x}}^{(t)}$ over the noisy image, and is empirically selected as 0.1, to achieve better denoising effect in follow-up iterations.
\begin{algorithm}[h]
	\caption{GSR Image Denoising with improved patch search scheme}
	\label{alg:algorithm3}
	\KwIn{The noisy image $\mathbf{y}\in \mathfrak{R}^{\sqrt{N}\times\sqrt{N}}$;
    
    \qquad \quad The white noise standard deviation, $\sigma$
    
    \qquad \quad The patch size $b$
    
    \qquad \quad The number of similar patches to extract 
    $c$
    
    \qquad \quad The learning rate $\mu$
    
    \qquad \quad The maximum number of iteration $max\_iter$
    
    \qquad \quad 

  }
    
	\KwOut{The denoised image, $\mathbf{x} \in\mathfrak{R}^{\sqrt{N}\times\sqrt{N}}$.}  
	\BlankLine

   $m \leftarrow \mathbf{\min}(b, c)$

    \For{$t=1$ to $max\_iter$}{

        \uIf{$t=1$}{\
            $\hat{\mathbf{x}}^{(t)} \leftarrow \mathbf{y},  \hat{\sigma}^{(t)} \leftarrow \sigma$
            
        }
        \Else{
        $\hat{\mathbf{x}}^{(t)} \leftarrow \mathbf{y} + \mu (\mathbf{y} - \hat{\mathbf{x}}^{(t-1)}),  \hat{\sigma}^{(t)} \leftarrow \sqrt{\sigma^2 - \|\mathbf{y-\hat{\mathbf{x}}^{(t)}}\|_2^2/N}$

        }
        \%search similar patches and perform grouping on $\hat{\mathbf{{x}}}^{(t)}$\%
        
        $\mathbf{x_{G}} \leftarrow \{\mathbf{x_{G_1}}, \cdots, \mathbf{x_{G_n}}\}$

        \For{k = 1 to n}{
        \%learn local dictionary $\mathbf{D_{G_k}}$\%
        
        \quad $\mathbf{x_{G_k}} \leftarrow \mathbf{U_kS_kV_k^T}, \mathbf{D_{G_k}} \leftarrow [\mathbf{u_{k,1}v_{k,1}^T},\cdots, \mathbf{u_{k,m}v_{k,m}^T}]$
        
        \%initialize $\mathbf{a_{G_k}}$ by full rank recovery \%
        
        $\mathbf{a_{G_k}} \leftarrow \mathbf{S_k}$
        
        }
        
        \%threshould coefficient vector $\mathbf{a_{G}}$\%
        
        $T \leftarrow 0$
        
        $rec\_loss \leftarrow 0$
        
        $prev\_rec\_loss \leftarrow 0$
        
        \While{
        $rec\_loss\leq 0.73\hat{\sigma}^{(t)}$
        }{  
            
            $\mathbf{a_{G}'} \leftarrow Hard(\mathbf{a_{G}}, T)$
            
            \%put sparsely decoded patches back to image\%
            
            $\mathbf{\hat{x}}^{'(t)} \leftarrow \mathbf{D_G\circ a_G'}$
            
            $rec\_loss \leftarrow \frac{1}{N}\|\mathbf{\hat{x}}^{'(t)} - \mathbf{\hat{x}}^{(t)}\|_1 $
        
        \uIf{$|prev\_rec\_loss-rec\_loss|<0.001$}{$\textbf{Break}$}
        {
            $\textbf{end}$

        }
       
        $prev\_rec\_loss\leftarrow rec\_loss$
            
        $T \leftarrow T+1$
        }
        $\mathbf{\hat{x}}^{(t)} \leftarrow \mathbf{\hat{x}}^{'(t)} $
        
        $t\leftarrow t+1$

    }

	\Return{$\hat{\mathbf{x}}^{(max\_iter)}$}
\end{algorithm}
\subsection{GSR Denoising Step}
The overall denoising step is as as follows, the first step is patch grouping for each patch by searching on the noisy image,  so as to obtain $\{\mathbf{x_{G_1}}, \cdots, \mathbf{x_{G_n}}\}$. The second step is iterative optimization. Before doing optimization, the current noise level $\hat{\sigma}^{(t)}$ need to be estimated, in the first iteration, the image to be denoised is the original noisy image and $\hat{\sigma}^{(t)}$ is the already known standard deviation of the white noise, but for following up iterations, the noisy level need to be estimated. Then for each group $\mathbf{x_{G_k}}$, learn its adaptive dictionary $\mathbf{D_{G_k}}$ by SVD decomposition and its coefficient $\mathbf{a_{G_k}}$ by threshoulding untill its mean absolute reconstruction difference is very close to $0.73\hat{\sigma}^{(t)}$. The third step is to put all sparsely represented patch in each patch group back to the image, and average the intensity by the number of times each pixel is counted to reconstruct a partially denoised version of 
the noisy image. Then this partially denoised image superimposes the noisy image
as the image to be denoised in the next iteration, and search similar patches on the superimposed noisy image, and repeat the second and third step until maximum allowed iteration is reached. Algorithm 1 gives a complete description of the GSR image denoising process. 

\section{Experimental Results}
This section displays all experimental results bwteen proposed method and other state-of-the-art methods. Also, the efficacy of improved similar patch seraching secheme, which search on the gradually less noisy image is compared with  searching only on noisy image. Both quantitative and qualitative results are reported.
All the test images are shown below, with size $256\times 256$.

\begin{figure}[h]
    \centering
    \includegraphics[width=1\linewidth]{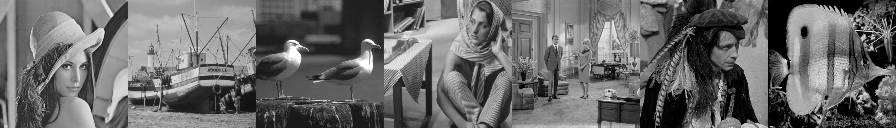}
    \caption{From left to right: \{Lena, Boat, Bird, Brab, Couple, Man, Fish\}}
    \label{fig:mesh1}
\end{figure}
The evaluation metric is peak signal-to-noise ratio (PSNR): $$MSE = \frac{1}{N}\|\mathbf{x} - \hat{\mathbf{x}}\|_2^2$$
$$PSNR = 20\log_{10}(255) - 10\log_{10}(MSE)$$

The quantitative result is displayed at Table 2, white noise with level $\sigma\in\{10, 20, 30, 50, 100\}$ are exhibited, in each grid, the top left is the result for Ksvd [1], the top right is the
result for BM-3d [2], the bottom left is the result for GSR image denoising without similar patch search scheme improved, the bottom left is the result for GSR image denoising with similar patch search scheme improved, while the bottom right is the result for GSR image denoising with similar patch search scheme improved. As we can see from the table, GSR-image denoising almost beated all results of Ksvd [1] and BM-3d [2], the top two results are labeled as bold font, most of them are from GSR-image denoising results. Also, it is clear to see that with improved patch similarity search scheme, GSR-image denoising can always produce better results.
\begin{table}[htbp]
\tiny
\centering
\begin{tabular}{|c| c|c|| c|c|| c|c|| c|c|| c|c|| c|c|| c|c| p{1cm}p{1cm}p{1cm}p{1cm}p{1cm}p{1cm}p{1cm}p{1cm}p{1cm}p{1cm}p{1cm}p{1cm}p{1cm}p{1cm}|}
\hline
$\sigma$ & \multicolumn{2}{|c|}{Lena}&\multicolumn{2}{|c|}{Boat} &\multicolumn{2}{|c|}{Bird}&\multicolumn{2}{|c|}{Brab}  &\multicolumn{2}{|c|}{Couple}&\multicolumn{2}{|c|}{Man}&\multicolumn{2}{|c|}{Fish}\\ \hline

\multirow{ 2}{*}{10} 
& 34.50 & 34.68                   & 33.03 & \textbf{33.52} & 35.20        & \textbf{35.51} & 33.04 & 33.39                   & 32.67 &\textbf{32.89} & 31.71 & \textbf{31.96} & 30.52 & 30.90\\ \cline{2-15}

& \textbf{34.80} & \textbf{35.09} & 33.41 & \textbf{33.68} & \text{35.42} & \textbf{35.75} & \textbf{33.63} & \textbf{33.91} & 32.78 &\textbf{33.02} & 31.94 & \textbf{32.19} & \textbf{30.94} & \textbf{31.13}\\ \hline
\hline
\multirow{ 2}{*}{20} 
& 30.25 & 31.14                   & 29.25 & 29.70                   & 31.78 & \textbf{32.04} & 29.55 & 29.89                   & 28.41 &29.09              & 27.55 & \textbf{27.91} & 26.44 & 26.90\\ \cline{2-15}

& \textbf{31.31} & \textbf{31.78} & \textbf{30.01} & \textbf{30.23} & 32.01 & \textbf{32.49} & \textbf{30.21} & \textbf{30.61} & \textbf{29.12} &\textbf{29.59} & 27.89 & \textbf{28.42} & \textbf{27.01} & \textbf{27.32}\\ \hline
\hline
\multirow{ 2}{*}{30} 
& 28.89 & 29.11 & 27.50 & 27.67 & 29.70 & 29.99 & 27.45 & 27.75 & 26.34 &26.96 & 25.50 & 25.78 & 24.40 & 24.73\\ \cline{2-15}

& \textbf{29.44} & \textbf{29.91} & \textbf{28.09} & \textbf{28.43} & \textbf{30.21} & \textbf{30.62} & \textbf{28.14} & \textbf{28.72} & \textbf{27.23} & \textbf{27.78} & \textbf{26.08} & \textbf{26.40} & \textbf{24.87} & \textbf{25.24}\\ \hline
\hline

%\multirow{ 2}{*}{50} 
%& 26.40 & 26.56 & 24.89 & 25.18 & 26.63 & 26.98 & 24.79 & 25.07 & 24.12 &24.46 & 23.00 & 23.12 & 21.90 & 22.04\\ \cline{2-15}

%& \textbf{27.01} & \textbf{27.50} & \textbf{25.70} & \textbf{26.21} & \textnf{27.21} & \textbf{27.56} & \textbf{25.94} & \textbf{26.48} & \textbf{25.01} &\textbf{25.60} & \textbf{23.13} & \textbf{23.77} & \textbf{22.32} & \textbf{22.71}\\ \hline
%\hline

\multirow{ 2}{*}{100} 
& 22.00 & 22.21 & 21.01 & 21.13 & 20.98 & 21.18 & 20.70 & 20.86 & 21.01 &21.13 & 18.53 & 18.77 & 17.76 & 17.99\\ \cline{2-15}

& \textbf{22.36} & \textbf{22.93} & \textbf{21.60} & \textbf{22.17} & \textbf{21.21} & \textbf{21.31} & \textbf{21.94} & \textbf{22.25} & \textbf{22.07} &\textbf{22.40} & \textbf{18.89} & \textbf{19.44} & \textbf{18.20} & \textbf{18.84}\\ \hline
\end{tabular}
\caption{The quantitative results for image denoising by using different algorithms}
\label{table2}
\end{table}

 \begin{figure}[h]
    \centering
    \includegraphics[width=1\linewidth]{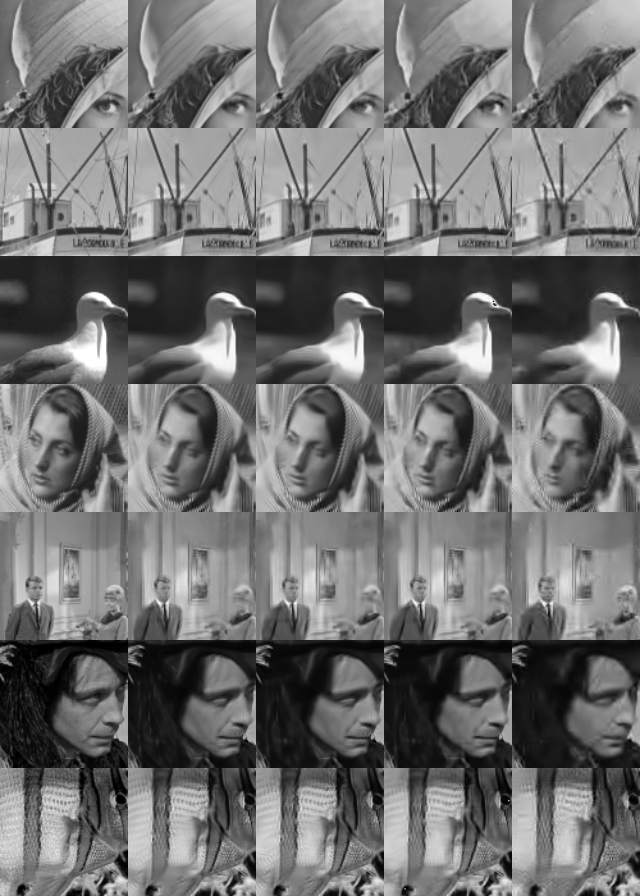}
    \caption{Denoised image when $\sigma$ = 20, Left to right: original, GSR(improved search scheme), GSR, BM-3d, Ksvd}

    \label{fig:mesh2}
\end{figure}

In Figure 3, qualitative results are displayed for all test images, GSR with improved search scheme preserves major parts of local texture while smoothing the flat area of original image, and this helps produce the most natural presentations. GSR lost a little bit local texture and looks over-smoothed because using not well matched groups tend to average out the local texture. BM-3d preserves most local textures but also produces lots of artifacts in the flat area or the original image. Ksvd produces the worst denosing effect with noticeable artifacts and damaged local textures.

\begin{figure}[h]
    \includegraphics[width=0.5\linewidth]{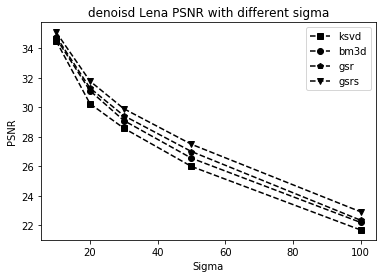}
    \includegraphics[width=0.5\linewidth]{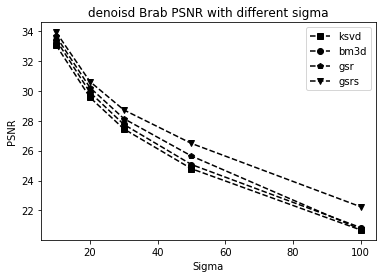}
    \caption{Denoised image when $\sigma = \{10, 20, 30, 50, 100\}$}

    \label{fig:mesh2}
\end{figure}

\begin{figure}[h]
    \includegraphics[width=0.5\linewidth]{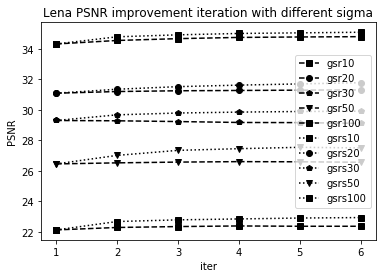}
    \includegraphics[width=0.5\linewidth]{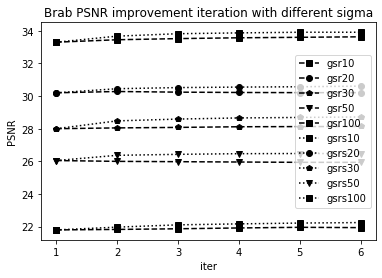}
    \caption{PSNR improvement over iterations}

    \label{fig:mesh2}
\end{figure}

Better similar patch searching scheme significantly helps denoising quality. In Figure 4, it is clear to see that at all noise levels, GSR image denoising with improved search scheme (GSRS) beats GSR and other two methods on test images Lena and Brab. Also, the advantage is clearly exhibited in iterative denoising process, as shown in Figure 5, better searching scheme  significantly boost denoising quality in the first few iterations, while the iterative improvement for naive similar patch search method only helps little.

\section{Conclusion}
This paper proposes a Group Sparse Representation based method for image denoising, that is, solving sparse coding problems for each similar patch group and iterate this process on more and more better quailty image. This method also naturally incorporate with a better patch search scheme, that is, searching on more and more better quality image. Finally this method is evaluated against by two state-of-the-art algorithms Ksvd and BM-3d, both qualitative and quantitative results are reported, and shows superior performance. In addition, improved similar patch search scheme is also evaluated by comparing with naive search scheme, it helps reaching better denoising quality both visually and statistically.

% {
% \bibliographystyle{ieee_fullname}
% \bibliography{egbib}
% }

\newpage
%\section*{References}

{\small
\bibliographystyle{ieee_fullname}
\bibliography{egbib}
}

%\small{
%[1] Elad, M., & Aharon, M. (2006). Image denoising via sparse and redundant representations over learned dictionaries. IEEE Transactions on Image processing, 15(12), 3736-3745.

%[2] Dabov, K., Foi, A., Katkovnik, V., & Egiazarian, K. (2007). Image denoising by sparse 3-D transform-domain collaborative filtering. IEEE Transactions on image processing, 16(8), 2080-2095.

%[3] Zhang, J., Zhao, D., & Gao, W. (2014). Group-based sparse representation for image restoration. IEEE Transactions on Image Processing, 23(8), 3336-3351.

%[4] Goldstein, T., & Osher, S. (2009). The split Bregman method for L1-regularized problems. SIAM journal on imaging sciences, 2(2), 323-343.
\end{document}